\documentclass[10pt, conference]{IEEEtran}

\usepackage{multirow}
\usepackage[pdftex]{graphicx}
\usepackage[cmex10]{amsmath}
\usepackage[tight,footnotesize]{subfigure}

\begin{document}

\title{An Evaluation of Deep CNN Baselines for Scene-Independent Person Re-Identification}

\author{\IEEEauthorblockN{Paul Marchwica, Michael Jamieson, Parthipan Siva}
\IEEEauthorblockA{Senstar Corporation\\
Waterloo, Canada\\
\{Paul.Marchwica, Mike.Jamieson, Parthipan.Siva\}@aimetis.com}
}

\maketitle

\newcommand{\LargeTrain}{$LargeSetTrain$}
\newcommand{\LargeTrainSize}[1]{\textit{LargeSetTrainSize}$_{#1}$}
\newcommand{\Market}{\textit{Market1501}}
\newcommand{\MarketTrain}{\textit{Market1501Train}}
\newcommand{\MarketTest}{\textit{Market1501Test}}
\newcommand{\OfficeBuilding}{\textit{OfficeBuilding}}

\begin{abstract}
In recent years, a variety of proposed methods based on deep convolutional neural networks (CNNs) have improved the state of the art for large-scale person re-identification (ReID). While a large number of optimizations and network improvements have been proposed, there has been relatively little evaluation of the influence of training data and baseline network architecture. In particular, it is usually assumed either that networks are trained on labeled data from the deployment location (scene-dependent), or else adapted with unlabeled data, both of which complicate system deployment. In this paper, we investigate the feasibility of achieving scene-independent person ReID by forming a large composite dataset for training. We present an in-depth comparison of several CNN baseline architectures for both scene-dependent and scene-independent ReID, across a range of training dataset sizes. We show that scene-independent ReID can produce leading-edge results, competitive with unsupervised domain adaption techniques. Finally, we introduce a new dataset for comparing within-camera and across-camera person ReID.
\end{abstract}


\section{Introduction}

Large public facilities, such as airports, host thousands of individuals who are observed by thousands of security cameras, that generate an overwhelming quantity of surveillance footage. Sufficiently accurate person re-identification (ReID) can greatly improve situational awareness by allowing operators to track people by appearance across the entire space. This ability could be used to find a lost child, quickly determine where a suspect has been, or help understand how people make use of the facility. 

As the number of potential applications have become apparent, ReID has received a great deal of attention in the literature \cite{Liu16,Liu17,Varior16,Karanam16,zheng2016person}. In recent years, methods based on deep convolutional neural networks (CNNs) \cite{ZhangXHL17, Zhong217, HermansBL17} have shown very promising results on some large datasets such as \Market{} \cite{market1501Dataset}. However, existing studies, including review papers \cite{zheng2016person,Karanam16}, do not fully evaluate some of the more practical needs of a person re-identification system, such as scene-independence and input resolution.

\begin{figure}[!h]
\begin{center}
\includegraphics[width=0.8\linewidth]{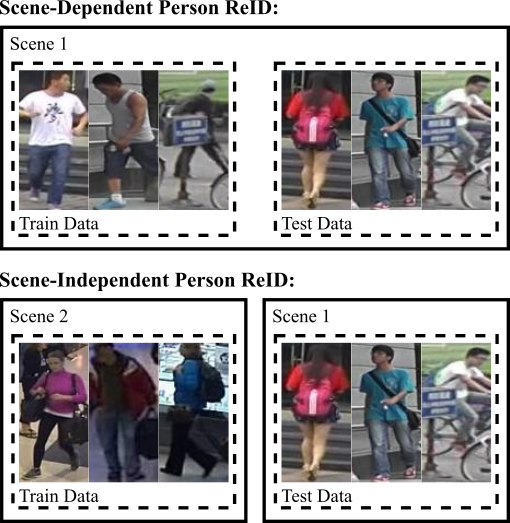}
\caption{Scene-dependent vs scene-independent person re-identification scenarios. In the well-studied scene-dependent scenario \cite{zheng2016person}, the training and test sets share camera views but not individuals. The scene-independent scenario is more pracitcal but also more challenging: there is no overlap in camera view or individuals between training and testing data. Here, Scene 1 has images from the \Market{} dataset \cite{market1501Dataset} while Scene 2 images come from the \textit{Airport} dataset\cite{Karanam16}. \label{fig:depvsindep}}
\end{center}
\end{figure}

A practical, working ReID system should not depend on labeled training data from its deployed environment. Manually collecting labeled correspondences for each camera at each prospective deployment site is too complex and expensive to be feasible. Even if some labeled data \emph{could} be collected, it probably would not capture the evolution of viewing conditions over the course of a day, let alone the years that a system would operate. Researchers understandably do not report `person-dependent' ReID accuracy where the test images are of the same individuals as the training images; it would be unrealistic to expect the system to have prior labeled data for each individual encountered. However, all promising deep ReID papers only report the \emph{scene-dependent} accuracy where test data is from the same environment as the training data \cite{ZhangXHL17, Zhong217, HermansBL17} (fig.~\ref{fig:depvsindep}). The current literature almost never evaluates the more realistic \emph{scene-independent} scenario where the training and test sets have no overlap in view, environment, or subjects. Even the few works \cite{crossData2015} that look at scene-dependent scenario do not report baseline method performances.

There are two complementary approaches to dealing with the lack of labeled data at the deployment site: achieving scene-independence by training on a dataset of sufficient size and variety, and achieving scene-independence by using unlabeled data to adapt to the deployed scene. If a large and truly representative dataset is available, then methods trained on it could potentially work on any new scene. However, no such dataset exists at the moment and Geng \emph{et al.} \cite{GengWXT16} argue that creating such a dataset would be too costly. Accordingly, there is a focus on domain adaptation techniques using unlabeled data \cite{Zhou217,FanZY17,Deng17} from a deployment environment. Unsupervised domain adaptation is an important area of research that promises improved accuracy, but comes at the expense of a much more complex deployment model. Even though such domain transfer approaches can benefit from using a scene-independent network as a starting point, the accuracy of baseline methods for scene-independent person ReID is not fully evaluated \cite{Zhou217,FanZY17,Deng17}.

Surveillance systems often rely on views from wide-angle cameras with many people in the scene, some of whom can be quite small in terms of pixels. ReID methods are rarely evaluated at the lower resolutions at which people can be reliably detected \cite{pedDetectionEval}. However, a deployed ReID system that ignores low-resolution images of people may miss a large number of useful matches. For practical surveillance applications, it is important to evaluate ReID methods across a range of input resolutions.

In surveillance, there are two distinctive uses of person ReID: finding a person across all cameras in a facility (across-camera person ReID) and finding all appearances of a person in the same camera view (within-camera person ReID). Existing literature has focused on across-camera ReID \cite{zheng2016person,Karanam16} since the general problem is seen as more challenging. However, within-camera ReID across large time gaps can be challenging due to changes in lighting, person pose, and clothing (e.g. removing a jacket). Moreover, within-camera ReID has many surveillance applications, such as commercial and residential lobby cameras where timing of individual entrances and exits would be valuable. One factor in the lower prominence of within-camera ReID may be lack of existing datasets.

In this paper we address the lack of scene-independent person ReID evaluation and within-camera person ReID evaluation. We use baseline deep CNN person ReID methods to investigate the current state of scene-independent person ReID. We present a mechanism for creating a scene-independent testing scenario using multiple existing person ReID datasets. Using these combined datasets, we evaluate scene-independent person ReID and compare it to baseline scene-dependent person ReID, state-of-the-art scene-dependent person ReID results, and state-of-the-art domain adaptation methods. Furthermore, we investigate the effects of input resolution and deep CNN network architecture on both scene-dependent and scene-independent person ReID.

Finally, we introduce a new person re-identification dataset collected during a full working day at the entrance of an office building. We use this dataset to compare the difficulty of the within-camera and across-camera modes of scene-independent ReID.

\section{Evaluation Setup}

\begin{figure}[!h]
\begin{center}
\includegraphics[width=0.95\linewidth]{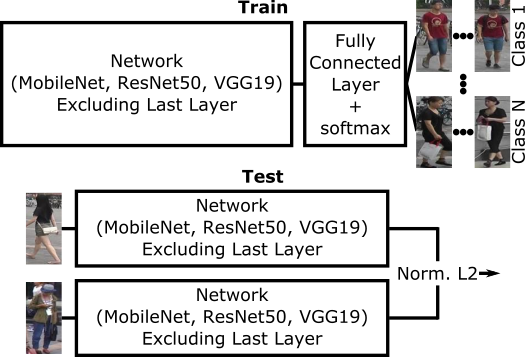}
\caption{ReID using Identification Networks. Deep CNNs are trained to identify people by setting each individual in the training data as a different class. The trained network, without the last fully connected layer, is then used at test time to compute features for two different images. A normalized Euclidean distance is then used to compare the features. \label{fig:idnetwork}}
\end{center}
\end{figure}

There are several ways to use deep convolutional neural networks (CNN) for person re-identification. Perhaps the simplest and most straightforward method is to train an identification network (fig.~\ref{fig:idnetwork}) and reuse the network up to the final feature layer to generate feature vectors at test time. That is, we train a network for classification, where each class is a unique person in the training set. Once trained, the layers associated with the final classification are dropped and the remaining network is used at test time to compute a feature vector from an input person image. The difference in person appearance between images can be quickly estimated from the corresponding feature vectors using a distance metric (usually normalized L2 distance). This identification network approach \cite{zheng2016person} is the most common and does not require formation of image pairs as in Siamese networks \cite{Zheng317} or triplet loss networks \cite{Liu16}. It also produces a compact feature vector which allows for larger-scale storage and appearance matching than the cross-difference method \cite{ahmed2015} that compares large feature sets within the network. As such, we will use this method for training our person re-identification networks.

We use the most broadly-accepted method (an identification network approach with normalized L2 distance) to investigate key aspects of person re-identification, as outlined below.

\subsection{Scene Independence}
\label{sec:SceneIndependenceTest}

Our main goal is to perform a baseline evaluation of scene-independent person re-identification. Since no study has looked at the baseline performance of this scenario, we first describe how we arrange a scene-independent person re-identification evaluation using the available datasets.

For scene-independent person ReID, there must be no overlap in camera view, environment, or people between the training and testing sets (fig.~\ref{fig:depvsindep}). This is not achievable using any single existing ReID dataset. However, using several datasets, we can achieve scene independence. We fix one dataset (\Market{}) as our testing dataset and train on a completely different dataset which has no overlap with the testing dataset. To allow for the best possible outcome, the training set should contain a multitude of scenes, environments, and people. This can be achieved by combining several person ReID datasets together into our training set (\LargeTrainSize{4}). In contrast, to test the performance of scene-dependent person ReID, we use the typical training (\MarketTrain{}) and testing (\MarketTest{}) subsets of \cite{market1501Dataset}.

Our main experiment will test scene-independent person re-identification by using \LargeTrainSize{4} to train our network and \MarketTest{} to test our network. We will compare this to scene-dependent person re-identification by training the same network on the \MarketTrain{} dataset and test on the \MarketTest{} set. Details of how we form \LargeTrainSize{4}, \MarketTrain{}, and \MarketTest{} sets can be found in section \ref{sec:datasets}.

\subsection{Within-camera vs Across-camera Re-Identification}

While most papers investigating person ReID focus on matching a pedestrian from one camera view to a different camera view \cite{market1501Dataset,zheng2016person} (across-camera ReID), no papers have looked at the problem of person re-identification in the same camera view but at different times (within-camera ReID). Within-camera ReID is arguably a simpler task as the camera geometry and camera parameters remain the same. However, environmental factors such as lighting, pedestrian pose, and clothing configuration can change. From an industry standpoint, this is still a desirable problem to solve for a variety of tasks. For example, given a video of a person entering a building, find when they left the building (assuming one entrance and exit, which is the case for many smaller stores and office buildings). 

Perhaps due to within-camera ReID being a simpler subproblem of across-camera ReID, or possibly because of the lack of a good dataset, the current literature does not directly evaluate within-camera ReID. In the case of scene-independent person re-identification, we would like to know if within-camera ReID performs better than across-camera ReID. To this end we introduce the \OfficeBuilding{} dataset.

\begin{figure}[!h]
\begin{center}
\includegraphics[width=0.8\linewidth]{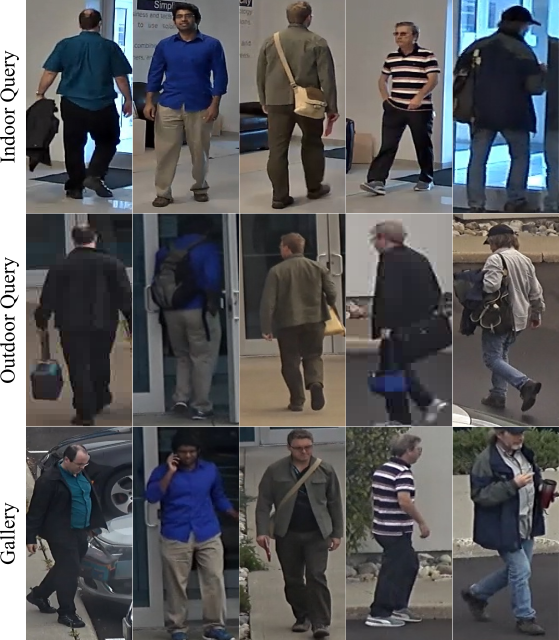}
\caption{\OfficeBuilding{} Dataset. There are query images from an indoor camera (for across-camera test) and outdoor camera (for within-camera test). The gallery is composed of people in the outdoor camera, excluding the query periods when a person first approached and entered the building. Since the data was collected during an entire day, people are sometimes compared front-view to back-view as well as with and without a jacket or backpack. \label{fig:officeBuildingDataset}}
\end{center}
\end{figure}

The \OfficeBuilding{} dataset contains two camera views (an outdoor and indoor view) of people entering and leaving a building during a single workday. Images from the outdoor camera are used for the gallery and the within-camera query. To ensure the within-camera query and the gallery do not have frames of an individual from the same time, we take all frames from a person entering the building for the first time (typically early morning) as the within-camera query image. People exiting the building and anyone seen re-entering the building on the outdoor camera are used as part of the gallery. All individuals seen in the indoor camera view are used for the across-camera view query image. Figure~\ref{fig:officeBuildingDataset} shows sample images from the \OfficeBuilding{} dataset.

\subsection{Network Architecture}

We wish to determine if a particular baseline CNN architecture is best suited for person re-identification. To this end, we test MobileNet \cite{MobileNetNetwork}, ResNet-50 \cite{ResNet50Network}, VGG19 \cite{VGG19Network}, and SqueezeNet \cite{SqueezeNetNetwork}. Another commonly used network architecture is Inception-v3 \cite{Inceptionv3Network}. However, due to the network depth and the amount of pooling employed, we are unable to test this network architecture at different input sizes without removing a significant number of layers.  For some of the more in-depth experiment such as dataset size (described in section \ref{sec:DatasetSize}), we test only the best performing network architectures.

\subsection{Input Size and Aspect Ratio}

Many existing works use images at 1:1 aspect ratio with resolutions of 224x224 or 227x227 \cite{zheng2016person}. This is most likely because the network architectures used have pre-trained weights using ImageNet data where images are rescaled and cropped to 224x224 or 227x227 resolution (224 being a practical number that is easy to divide and perform convolutions on). There are two important points to consider: 1) ImageNet resizing and cropping generally preserve aspect ratio \cite{VGG19Network} and 2) network architectures that use global pooling are somewhat robust to input image size, so long as the input size is big enough for all the pooling layers. Given that pedestrian detectors were shown to be effective for person heights as low as 50 pixels \cite{pedDetectionEval}, we investigate input sizes of 64x32, 128x64, and 256x128.

For comparisons between 2:1 and 1:1 aspect ratios we stretch the 256x128 input size to 256x256. We choose 256x256 for comparison because it is close in size to the typically used 224x224 resolution and is simply a stretch from our 256x128 resolution. If we used 224x224 resolution to compare to 256x128 resolution, we could not isolate the influence of aspect ratio from the scale change.

\subsection{Dataset Size}
\label{sec:DatasetSize}

For scene-independent person ReID, the training set should be large and varied enough to span the range of conditions likely to appear in the test set. To test how much scene-independent ReID accuracy depends on size and composition of the training set, we consider a series of six progressively larger training sets, \LargeTrainSize{1\ldots 6}. The details of \LargeTrainSize{1\ldots 6} can be found in section \ref{sec:datasets}.

\subsection{Network Initialization}

All existing approaches to person ReID \cite{zheng2016person} initialize the network using weights trained on ImageNet. However, no papers have tested whether random initialization can achieve similar results. In this study, we compare training these networks initialized both randomly and with ImageNet weights.

\section{Evaluation Implementation}

\subsection{Datasets} 
\label{sec:datasets}
We select a number of commonly used ReID datasets for use in our evaluation procedures. The primary datasets used in this study are: \Market{} \cite{market1501Dataset}, \textit{Airport} \cite{Karanam16}, \textit{DukeMTMC4ReID} \cite{dukemtmc4reidDataset}, \textit{LSPS} \cite{LargeScalePersonSearchDataset}, and \textit{CUHK03} \cite{CUHK03Dataset}.

The \Market{} dataset consists of 32,668 images and 1,501 identities across six cameras.
The \textit{Airport} dataset consists of 39,902 images and 9,651 identities across six cameras.
The \textit{DukeMTMC4ReID} dataset consists of 46,261 images and 1,852 identities across eight cameras.
The \textit{LSPS} dataset consists of 18,184 images and 8,432 identities.
The \textit{CUHK03} dataset consists of 13,164 images and 1,360 identities across six cameras.

Our newly introduced dataset, \OfficeBuilding{}, is also used in testing. This dataset includes 37 identities and 1,279 images captured across two cameras. It contains 21 identities and 210 images which can be used for the across-camera query, as well as 34 identities and 340 images which can be used for the within-camera scenario.

We split \Market{} into a training \MarketTrain{} and test \MarketTest{} set. The split is the standard $751/750$ split used in other published works \cite{market1501Dataset}.

For all datasets we keep $10$ images per person and ensure images are selected from all camera views the person appears in. If a person appears in more than $10$ camera views, one image per camera view is kept. This balances the dataset as some datasets have a large number of images for each person compared to others. This sampling approach is motivated by the work of Xiao et al. \cite{Xiao2016CVPR}, who hypothesize that failing to balance the data from multiple sources can lead to overfitting from smaller datasets. For each dataset, $90\%$ of the unique person IDs are used for ReID training, while the remaining $10\%$ are withheld for testing. Since we are training an ID network, each unique person is treated as a single class. For each person in the training sets, the images are split into training and validation sets at a $70-30$ random split. 

In particular, we seek to evaluate how the size and composition of the training set affects the scene-independent accuracy of a person ReID network. To this end, we have created a number of composite datasets. Table~\ref{TAB:DatasetComposition} gives an outline of the dataset groupings used and their composite names.

\LargeTrainSize{4}, which comprises \textit{DukeMTMC4ReID}, \textit{Airport}, \textit{CUHK03}, and \textit{LSPS}, is our primary `large set' used in training models. \LargeTrainSize{5} and \LargeTrainSize{6} are marked as using `Full Data', which indicates these use the entire datasets (no train/test split) with $10$ images per person. In order to have a more thorough test of the effects of added data, \LargeTrainSize{6} contains additional datasets, including: \textit{3DPES}\cite{3DPESDataset}, \textit{CUHK01}\cite{CUHK01Dataset}, \textit{iLIDS-VID}\cite{ilidsDataset}, \textit{PRID 2011}\cite{PRID2011Dataset}, \textit{Shinpuhkan2014}\cite{Shinpuhkan2014Dataset}, \textit{underGround Re-Identification (GRID)}\cite{GRIDDataset}, \textit{VIPeR}\cite{ViperDataset} and \textit{WARD}\cite{WARDDataset}.

\begin{table}[t]
	\caption{Contents of Composite Datasets}
	\label{TAB:DatasetComposition}
	\begin{center}
		\setlength{\tabcolsep}{0.5mm}
		\scriptsize{
			\begin{tabular}{|c|c|c|c|c|c|c|}
				\hline
				\multirow{2}{*}{Training Set}&\multirow{2}{*}{Full$^*$}&\multirow{1}{*}{\textit{DukeMTMC4ReID}} &\multirow{1}{*}{\textit{Airport}}&\multirow{1}{*}{\textit{CUHK03}}&\multirow{1}{*}{\textit{LSPS}}&\multirow{2}{*}{Extra$^{**}$} \\
				&&\cite{dukemtmc4reidDataset}&\cite{Karanam16}&\cite{CUHK03Dataset}&\cite{LargeScalePersonSearchDataset}& \\
				\hline
				\LargeTrainSize{1} & N & Y & & & & \\
				\hline
				\LargeTrainSize{2} & N & Y & Y & & & \\
				\hline
				\LargeTrainSize{3} & N & Y & Y & Y & & \\
				\hline
				\LargeTrainSize{4} & N & Y & Y & Y & Y & \\
				\hline
				\LargeTrainSize{5} & Y & Y & Y & Y & Y & \\
				\hline
				\LargeTrainSize{6} & Y & Y & Y & Y & Y & Y \\
				\hline
			\end{tabular}
		}
	\end{center}
$^*$ Full Data does not reserve any individuals for testing\\
$^{**}$ Extra=\{\textit{3DPES}\cite{3DPESDataset}, \textit{CUHK01}\cite{CUHK01Dataset}, \textit{iLIDS-VID}\cite{ilidsDataset}, \textit{PRID 2011}\cite{PRID2011Dataset}, \textit{Shinpuhkan2014}\cite{Shinpuhkan2014Dataset}, \textit{GRID}\cite{GRIDDataset}, \textit{VIPeR}\cite{ViperDataset}, \textit{WARD}\cite{WARDDataset}\}
\vspace*{-2mm}
\end{table}

\subsection{Architecture} 
\label{sec:arch}

We use the following networks in our evaluation: MobileNet, ResNet50, VGG19, and SqueezeNet. At training time, for all networks except SqueezeNet, we remove the final fully-connected layer and replace it with a new fully-connected layer with output dimension based on the number of classes (individuals) in our training data. At test time, the output of the layer before the last fully-connected layer is used as our feature. We use normalized L2 distance as our distance metric.

SqueezeNet does not use a fully-connected layer for classification. Instead, it uses a 1x1 convolution layer with as many channels as there are classes, followed by global pooling before the softmax layer. During training, we set the number of channels in this 1x1 convolution layer to the number of classes (individuals) in our training data. At test time, the output of the layer before the 1x1 convolution layer is fed into a global average pooling layer and the output of the pooling layer is used as our feature.

Since we minimized structural alterations of existing network architectures for obtaining features, we end up with different feature dimensionality for each architecture. The dimensions used are: MobileNet-1024D, ResNet50-2048D, VGG19-4096D, and SqueezeNet-512D.

\subsection{Solver And Data Augmentation}

When training the network, we used stochastic gradient descent (SGD) as our optimizer with a learning rate of 0.001, and default decay/momentum parameters. A batch size of 6 was used for all training. Unless otherwise stated, all networks were trained for 160 epochs, where each epoch is one iteration through the entire training data.

Since our focus is a baseline evaluation, we do not introduce dropout or other fine-tuning techniques. Furthermore, we keep data augmentation simple and use only horizontal flips and random cropping.

\subsection{Hardware/Software Details}

All networks are trained and evaluated using TensorFlow (1.3.0) and Keras (2.0.9) on machines with NVIDIA GeForce GPUs (GTX 1060 and 1080).

\section{Results}

For the purposes of this evaluation, we limit our investigation to single query mode. We report the typical Rank-1 and Rank-5 accuracy based on the cumulative match curve (CMC) and the mean average precision (mAP) \cite{ViperDataset}. For a consistent evaluation, we use the code provided by \cite{zheng2016person,zhengLatestResult}.

\subsection{Architecture and Input Size}

\begin{table*}[t]
\caption{\MarketTest{} Results}
\label{TAB:MainResults}
\begin{center}
\setlength{\tabcolsep}{0.5mm}
\scriptsize{
\begin{tabular}{|c|c|c|c|c|c|c|c|c|c|c|c|c|c|}
\hline
\multirow{2}{*}{Input Size}&\multirow{2}{*}{Training Set}&\multicolumn{3}{c|}{MobileNet}&\multicolumn{3}{c|}{ResNet50}&\multicolumn{3}{c|}{VGG19}&\multicolumn{3}{c|}{SqueezeNet} \\
\cline{3-14}
&&Rank-1&Rank-5&mAP&Rank-1&Rank-5&mAP&Rank-1&Rank-5&mAP&Rank-1&Rank-5&mAP \\
 \hline
 \hline
 256x128 & \MarketTrain & \textbf{75.6\%} & 88.9\% & 0.49 & 72.5\% & 87.9\% & 0.48 & 65.4\% & 81.9\% & 0.38 & 59.9\% & 77.5\% & 0.33 \\
 \hline
 128x64 & \MarketTrain & 73.6\% & 88.8\% & 0.45 & \textbf{74.2\%} & 87.6\% & 0.48 & 66.0\% & 82.3\% & 0.40 & 58.9\% & 77.8\% & 0.34 \\
 \hline
 64x32 & \MarketTrain & 68.1\% & 85.4\% & 0.41 & \textbf{69.1\%} & 86.2\% & 0.44 & 58.5\% & 78.1\% & 0.34 & 49.2\% & 71.4\% & 0.30 \\
 \hline
  \hline
 256x128 & \LargeTrainSize{4} & \textbf{46.9\%} & 64.5\% & 0.21 & 43.5\% & 61.5\% & 0.19 & 32.2\% & 50.2\% & 0.12 & 29.6\% & 46.8\% & 0.11 \\
 \hline
 128x64 & \LargeTrainSize{4} & \textbf{49.3\%} & 68.2\% & 0.23 & 45.6\% & 64.2\% & 0.21 & 36.0\% & 54.5\% & 0.15 & 31.0\% & 49.4\% & 0.12 \\
 \hline
 64x32 & \LargeTrainSize{4} & 40.7\% & 60.1\% & 0.18 & \textbf{44.1\%} & 62.4\% & 0.19 & 32.3\% & 50.7\% & 0.13 & 26.6\% & 44.9\% & 0.10 \\
 \hline
\end{tabular}
}
\end{center}
\vspace*{-3mm}
\end{table*}

Table \ref{TAB:MainResults} summarizes our main finding. We can see that for the scene-dependent case, MobileNet and ResNet50 have the best performance of the networks tested by a fair margin ($\sim30\%$). MobileNet and ResNet50 also have the best performance for the scene-independent case where the training set is different than the test set

With regards to input size, 128x64 has fairly consistent performance across all networks. 256x128 resolution has slightly better performance in some cases, such as MobileNet trained on \MarketTrain{} data. In this scene-dependent scenario, the 256x128 resolution has $2\%$ better Rank-1 performance than the 128x64 resolution. However, compared to the 256x128, the 128x64 input resolution has $2.4\%$ better performance on MobileNet for the more challenging scene-independent case. The 64x32 resolution has the lowest accuracy, but it could still generate useful results for people who are more distant from the camera.

Overall, MobileNet with the 128x64 input resolution has the best performance for the scene-independent test scenario, followed by ResNet50 with the same resolution. As a result, the remaining investigations focus on the MobileNet and ResNet50 network architectures with an input size of 128x64.

\begin{figure*}[!tb]
\begin{center}
\subfigure[]{
\includegraphics[width=0.4\linewidth]{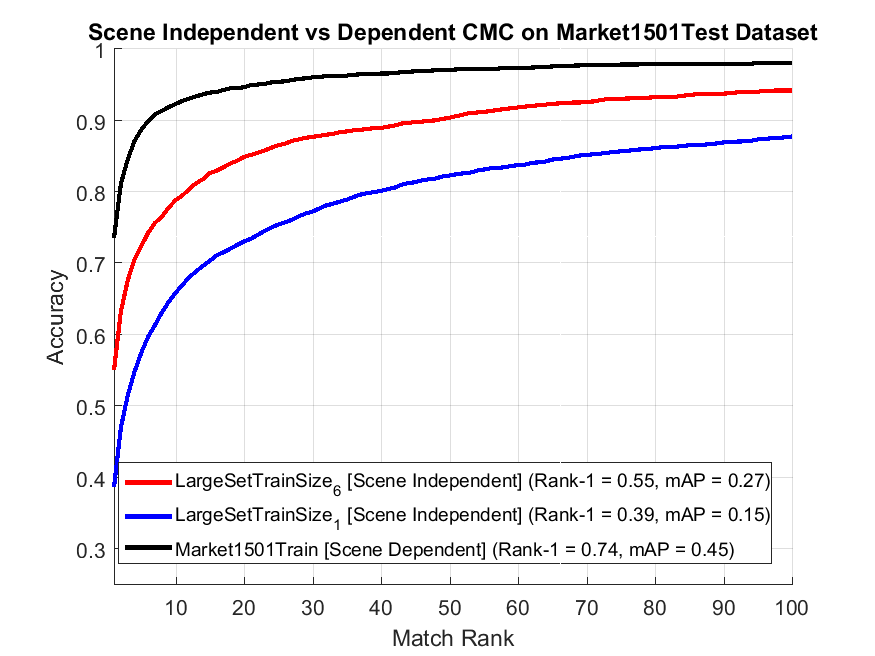} \label{subfig:marketcmc}
}
\subfigure[]{
\includegraphics[width=0.4\linewidth]{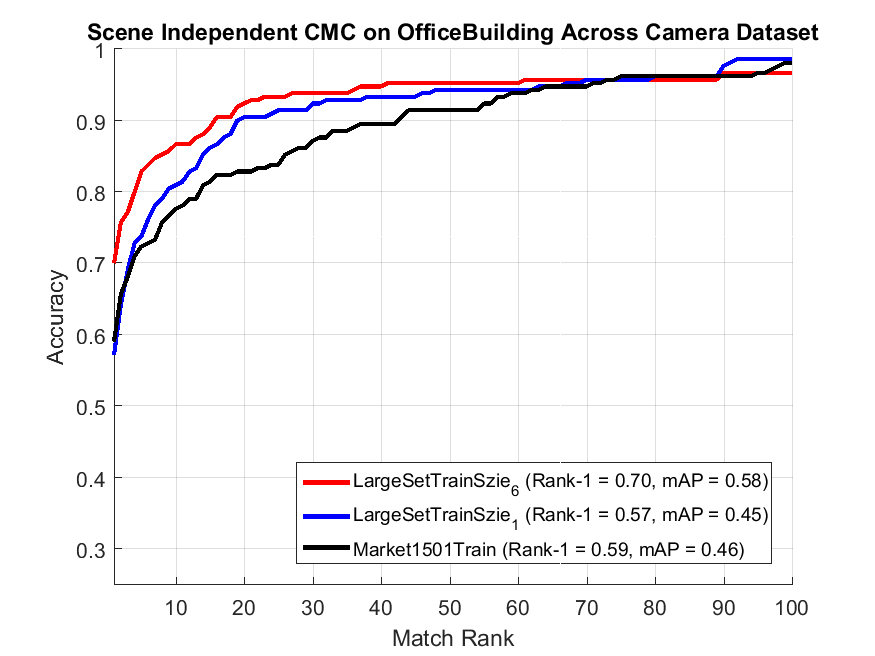} \label{subfig:officebuildingcmc}
}
\caption{Cumulative Matching Curve (CMC) on the \MarketTest{} and \OfficeBuilding{} datasets using MobileNet at 128x64 input resolution. We see that scene-dependent scenario has much better performance (trained on \MarketTrain{} and tested on \MarketTest{}) than scene-independent scenario. For the scene-independent scenario, performance is controlled by the training dataset size (\LargeTrainSize{1} is smaller than \LargeTrainSize{6}). \label{fig:cmc}}
\end{center}
\end{figure*}

\subsection{Scene Independence and Dataset Size}
\label{SEC:ResultsSceneIndDatasetSize}

From Table \ref{TAB:MainResults} and fig.~\ref{subfig:marketcmc} we can see that scene-independent person re-identification results are consistently much lower than scene-dependent person re-identification regardless of the network or the input size used. There is a staggering $\sim30\%$ Rank-1 accuracy difference. This is to be expected because even though \LargeTrainSize{4} has combined several datasets together, it still lacks the diversity needed to represent the \MarketTest{} dataset.

To see the effect of training set size on the scene-independent re-identification scenario, we trained MobileNet and ResNet50 across a range of training set sizes, with the results reported in Table~\ref{TAB:DatasetSizeResults} and fig.~\ref{fig:cmc}. As the training dataset size increases, the accuracy of both the MobileNet and ResNet50 models increase. We can also see that the ResNet50 model is consistently lagging the MobileNet performance by $\sim 2 - 5 \%$. Overall, using our largest training set and MobileNet, we can achieve a Rank-1 accuracy of $55.1\%$ for the scene-independent person re-identification scenario. Though this is still $\sim20\%$ below the results in the scene-dependent scenario, these results are encouraging. The trend suggests that if we can build a large enough training set, we can train a model that can handle scene-independent person re-identification.

\begin{table}[t]
\caption{Results For Different Dataset Size}
\label{TAB:DatasetSizeResults}
\begin{center}
\setlength{\tabcolsep}{0.5mm}
\scriptsize{
\begin{tabular}{|c|r|r|c|c|c|c|c|c|}
\hline
\multirow{2}{*}{Training Set}&\multicolumn{1}{c|}{\#} &\multicolumn{1}{c|}{\#}&\multicolumn{3}{c|}{MobileNet}&\multicolumn{3}{c|}{ResNet50} \\
\cline{4-9}
&People&Images&Rank-1&Rank-5&mAP&Rank-1&Rank-5&mAP \\
 \hline
\LargeTrainSize{1} & 1,272 & 7,979 & \textbf{38.7\%} & 57.5\% & 0.15 & 36.6\% & 55.2\% & 0.14 \\
\hline
\LargeTrainSize{2} & 2,515 & 12,851 & \textbf{41.6\%} & 61.2\% & 0.16 & 38.2\% & 56.1\% & 0.15 \\
\hline
\LargeTrainSize{3} & 3,836 & 21,729 & \textbf{47.6\%} & 66.5\% & 0.21 & 42.3\% & 60.6\% & 0.19 \\
\hline
\LargeTrainSize{4} & 14,577 & 46,423 & \textbf{49.3\%} & 68.2\% & 0.23 & 45.6\% & 64.2\% & 0.21 \\
\hline
\LargeTrainSize{5} & 16,195 & 68,763 & \textbf{54.8\%} & 71.6\% & 0.27 & 51.9\% & 69.4\% & 0.25 \\
\hline
\LargeTrainSize{6} & 19,092 & 84,080 & \textbf{55.1\%} & 72.5\% & 0.27 & 52.2\% & 69.1\% & 0.26 \\
 \hline
\end{tabular}
}
\end{center}
\end{table}

\subsection{Comparison To The State-Of-The-Art}

Overall, our best results are for MobileNet with an input size of 128x64. For this configuration, the Rank-1 accuracy is $73.6\%$ for the scene-dependent scenario (Table \ref{TAB:MainResults}) and $55.1\%$ for the scene-independent scenario (Table \ref{TAB:DatasetSizeResults}). Please note that the addition of dropout, data augmentation such as rotations, etc. was not the focus of this study and could potentially be used to increase these accuracies \cite{zhengLatestResult}. Regardless, if we compare these baseline results to published methods (Table \ref{TAB:StateOfArtResults}), we find that they fall in the middle of the pack. The scene-independent Rank-1 result of $55.1\%$ even beats some of the scene-dependent results as recent as 2016.

\begin{table}[t]
\caption{State-of-the-Art Comparisons for Scene-Dependent ReID}
\label{TAB:StateOfArtResults}
\begin{center}
\setlength{\tabcolsep}{0.5mm}
\scriptsize{
\begin{tabular}{|c|c|c|c|c|c|c|c|c|c|c|}
\hline
Alg. & \cite{Wu16} & \cite{Sh16} & \cite{Liu16} & \cite{Jose16} & \cite{Karanam16} & \cite{Martinel16} & \cite{Wu216} & \cite{Liu17} & \cite{Chen16} & \cite{zhang16} \\
\hline
Rank-1 & 37.2 & 39.4 & 45.1 & 45.2 & 46.5 & 47.9 & 48.2 & 48.2 & 51.9 & 55.4 \\
\hline
\hline
Alg. & \cite{Ustinova16} & \cite{Varior16} & \cite{Varior216} & \cite{Ustinova17} & \cite{Zhou17} & \cite{Chen17} & \textbf{RN L} & \textbf{MN M} & \cite{Lin17} & \cite{Barbosa17} \\
\hline
Rank-1 & 59.5 & 61.6 & 65.9 & 66.4 & 70.7 & 71.8 & 72.5 & 73.6 & 73.8 & 73.9 \\
\hline
\hline
Alg. & \textbf{RN M} & \textbf{MN L} & \cite{Zheng17} & \cite{zhao17} & \cite{Zhong17} & \cite{Zheng217} & \cite{Zheng317} & \cite{Li17} & \cite{Zhao217} & \cite{Bai17} \\
\hline
Rank-1 & 74.2 & 75.6 & 76.8 & 76.9 & 77.1 & 79.3 & 79.5 & 80.3 & 81.0 & 82.2 \\
\hline
\hline
Alg. & \cite{Yu17} & \cite{Sun17} & \cite{Zheng417} & \cite{GengWXT16} & \cite{Li217} & \cite{Su17} & \cite{Lin217} & \cite{HermansBL17} & \cite{Zhong217} & \cite{ZhangXHL17} \\
\hline
Rank-1 & 82.3 & 82.3 & 82.8 & 83.7 & 83.9 & 84.1 & 84.3 & 84.9 & 87.1 & 87.7 \\
\hline
\end{tabular}
}
\vspace{0.2cm}

\textbf{MN} MobileNet, \textbf{RN} ResNet50 \\
\textbf{M} 128x64 resolution, \textbf{L} 256x128 resolution
\end{center}
\end{table}

Interestingly, our scene-independent result is very competitive with unsupervised domain adaptation work (Table \ref{TAB:SOARUnsupDomainTransfer}). The unsupervised domain adaptation work is similar to our result in that it uses labeled data from other datasets. However, in addition to the scene-independent training data, it uses scene-dependent training data without ground truth annotation. That is, domain adaptation systems use the \MarketTrain{} dataset without ground truth label in their training process whereas our result does not use \MarketTrain{} data at all. The competitive nature of our result, even though we do not use scene-dependent training data, is explained by the large amount of training data we used by combining all other person re-identification datasets into a single dataset. Again, this shows that a large enough training dataset from multiple environments should be able to do scene-independent person re-identification without needing domain adaptation.

\begin{table}[t]
\caption{Unsupervised Domain Transfer}
\label{TAB:SOARUnsupDomainTransfer}
\begin{center}
\setlength{\tabcolsep}{0.5mm}
\scriptsize{
\begin{tabular}{|c|c|c|c|c|c|c|c|c|c|c|}
\cline{2-5}
\multicolumn{1}{c|}{} & \multicolumn{3}{c|}{Scene Dependent} & Scene Independent \\
\multicolumn{1}{c|}{} & \multicolumn{3}{c|}{\textit{Uses Unlabeled}} & \textit{Not Using} \\
\multicolumn{1}{c|}{} & \multicolumn{3}{c|}{\MarketTrain{}} & \MarketTrain{} \\
\hline
Alg. & \hspace{0.08cm} \cite{Zhou217} \hspace{0.08cm} & \hspace{0.08cm} \cite{FanZY17} \hspace{0.08cm} & \hspace{0.08cm} \cite{Deng17} \hspace{0.08cm} & \textbf{MobileNet 128x64} \\
\hline
Rank-1 & 40.9 & 44.7 & 57.7 & 55.1 \\
\hline
\end{tabular}
}
\end{center}
\vspace*{-5mm}
\end{table}

\subsection{Network Initialization}

The results thus far have been for networks initialized with ImageNet weights. This is a standard technique used in many different applications because it has been shown to boost performance in comparison to random initialization of the network. Table \ref{TAB:InitTypeResults} summarizes our result for random initialization vs ImageNet initialization. ImageNet initialization gives a boost of $10\%$ or more in comparison to random initialization for both scene-independent and scene-dependent scenarios.

The benefit of ImageNet initialization is a mixed blessing. While it gives a huge boost in performance, it discourages making any changes to existing network architectures that would require retraining on ImageNet data.

\begin{table}[t]
\caption{Results For Different Initialization}
\label{TAB:InitTypeResults}
\begin{center}
\setlength{\tabcolsep}{0.5mm}
\scriptsize{
\begin{tabular}{|c|c|c|c|c|c|c|c|}
\hline
\multirow{2}{*}{Training Set}&\multirow{2}{*}{Init. Type} &\multicolumn{3}{c|}{MobileNet}&\multicolumn{3}{c|}{ResNet50} \\
\cline{3-8}
&&Rank-1&Rank-5&mAP&Rank-1&Rank-5&mAP \\
\hline
\MarketTrain & ImageNet & 73.6\% & 88.8\% & 0.45 & 74.2\% & 87.6\% & 0.48 \\
\hline
\MarketTrain & Rand & 62.9\% & 82.3\% & 0.37 & 64.2\% & 81.9\% & 0.38 \\
\hline
\LargeTrainSize{4} & ImageNet & 49.3\% & 68.2\% & 0.23 & 45.6\% & 64.2\% & 0.21 \\
\hline
\LargeTrainSize{4} & Rand & 33.5\% & 52.8\% & 0.14 & 33.9\% & 53.3\% & 0.13 \\
\hline
\end{tabular}
}
\end{center}
\vspace*{-5mm}
\end{table}

\subsection{Aspect Ratio}

We look at the result of using a 2:1 versus a 1:1 aspect ratio input size in Table \ref{TAB:AspectRatioResults}. When comparing the 256x128 person size to the stretched version at 256x256, we see a lower performance for the stretched version. In particular, ResNet50 shows a nearly $7\%$ drop in performance. Since the 256x256 input also requires more computation than 256x128 input, the 2:1 aspect ratio is preferred. These results are consistent with the findings of \cite{YaoZZLT17}, who performed a similar test.

\begin{table}[t]
\caption{Results for Different Aspect Ratio}
\label{TAB:AspectRatioResults}
\begin{center}
\setlength{\tabcolsep}{0.5mm}
\scriptsize{
\begin{tabular}{|c|c|c|c|c|c|c|}
\hline
\multirow{2}{*}{Input Size}&\multicolumn{3}{c|}{MobileNet}&\multicolumn{3}{c|}{ResNet50} \\
\cline{2-7}
&Rank-1&Rank-5&mAP&Rank-1&Rank-5&mAP \\
\hline
256x128 & 75.6\% & 88.9\% & 0.49 & 72.5\% & 87.9\% & 0.48 \\
\hline
256x256 & 73.2\% & 87.5\% & 0.48 & 65.4\% & 83.3\% & 0.40 \\
\hline
\end{tabular}
}
\end{center}
\vspace*{-5mm}
\end{table}

\subsection{Within-camera vs Across-camera Re-identification}

\begin{figure}[!h]
\begin{center}
\includegraphics[width=\linewidth]{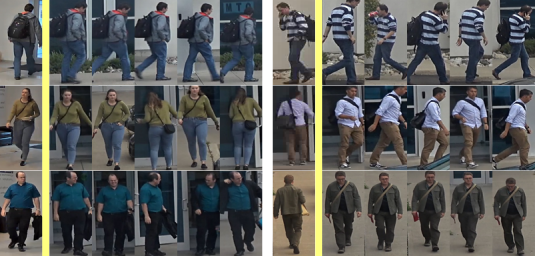}
\caption{Sample scene independent search results on the \OfficeBuilding{} dataset. Left: across-camera search results. Right: within-camera search result. \label{fig:officeBuildingSampleResults}}
\end{center}
\end{figure}

Using our \OfficeBuilding{} dataset, we test the accuracy of scene-independent person re-identification for within-camera and across-camera scenarios. We report the results using MobileNet trained on the \LargeTrainSize{6} and \MarketTrain{} in Table \ref{TAB:OfficeBuildingResults}. Some figures of the results are presented in fig.~\ref{fig:officeBuildingSampleResults}. Section \ref{SEC:ResultsSceneIndDatasetSize} illustrates that, once again, training on a larger, more varied dataset boosts performance. For the \OfficeBuilding{} dataset, there is a boost in Rank-1 accuracy of $\sim10\%$ when using a larger training set for both within and across-camera person re-identification.

When comparing within-camera to across-camera person re-identification, we find that there is a $\sim5\%$ difference. Meaning that matching across-camera is not much more difficult in this instance than matching within-camera. This is perhaps because individuals are matched across large time gaps, typically from morning to evening. Furthermore, we are searching using a person entering the building as the probe and the gallery contains mostly people exiting the building. As a result, we are mostly matching front view to rear view. 

\begin{figure}[!h]
\begin{center}
\includegraphics[width=\linewidth]{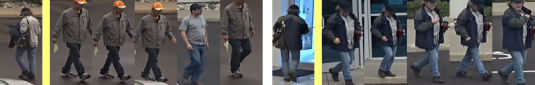}
\caption{Left: In an outdoor query image, the arriving subject has no coat, and the correct match is not found because the gallery has no image of the subject without a coat. Right: In an indoor query image, the subject is wearing a coat when exiting and the indoor-outdoor match quality is high. \label{fig:officeBuildingCoat}}
\end{center}
\end{figure}

\begin{figure}[!h]
\begin{center}
\includegraphics[width=\linewidth]{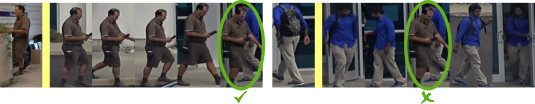}
\caption{A gallery image containing two people is only annotated as the foreground person (left query). Though it is also a reasonable match for the occluded person (right query), it is counted as incorrect. \label{fig:officeBuildingMultiplePeople}}
\end{center}
\end{figure}

The \OfficeBuilding{} dataset also contains some very challenging scenarios. For example, there are cases in the dataset where a person is not wearing a coat when entering the building but wearing a coat when exiting fig.~\ref{fig:officeBuildingCoat}. In another example, there are overlapping individuals but the ground truth annotation considers the person in the foreground as the true label fig.~\ref{fig:officeBuildingMultiplePeople}.

The scene-independent test result on the \MarketTest{} dataset ($55.1\%$ from Table \ref{TAB:DatasetSizeResults}) is much lower than the numbers reported on our \OfficeBuilding{} dataset because our data set is much simpler with only 37 unique individuals. This dataset's primary purpose is for comparison of within-camera and across-camera person re-identification.

\begin{table}[t]
\caption{\OfficeBuilding{} Results Using MobileNet}
\label{TAB:OfficeBuildingResults}
\begin{center}
\setlength{\tabcolsep}{0.5mm}
\scriptsize{
\begin{tabular}{|c|c|c|c|c|c|c|}
\hline
\multirow{2}{*}{Training Set}&\multicolumn{3}{c|}{across-camera}&\multicolumn{3}{c|}{within-camera} \\
\cline{2-7}
&Rank-1&Rank-5&mAP&Rank-1&Rank-5&mAP \\
 \hline
\MarketTrain & 59.1\% & 72.4\% & 0.46 & 64.7\% & 77.1\% & 0.58 \\
\hline
\LargeTrainSize{6} & 70.0\% & 82.9\% & 0.58 & 75.3\% & 84.4\% & 0.63 \\
 \hline
\end{tabular}
}
\end{center}
\end{table}

\section{Conclusion}

We performed an extensive evaluation of baseline deep convolution networks for person re-identification and found that on average MobileNet with input resolution of 128x64 is the best network for both scene-dependent and scene-independent person re-identification. Using MobileNet with 128x64 input, we can achieve $73.6\%$ for scene-dependent person re-identification on \MarketTest{}, which is near the middle of the range of reported results for this dataset.

Furthermore, MobileNet with 128x64 input resolution can perform well for scene-independent person re-identification when using a sufficiently large training dataset. While the result of the scene-independent test is $18.5\%$ lower than the scene-dependent scenario, it is still very competitive with unsupervised domain adaptation techniques which use unlabeled data from the \Market{} dataset during training.

We introduced a dataset to test scene-independent person re-identification when the probe is from the same camera as the gallery (within-camera person re-identification) and when the probe is from a different camera (across-camera person re-identification). Surprisingly, we find that both within-camera and across-camera results are very close ($\sim5\%$ difference). This indicates that even matching people in the same camera view across large time gaps could be as challenging as matching people across-camera views.

% Generated by IEEEtran.bst, version: 1.13 (2008/09/30)

\end{document}